%% file: autoped.tex
\documentclass[10pt,twocolumn,letterpaper]{article}

\usepackage{cvpr}
\usepackage{times}
\usepackage{epsfig}
\usepackage{graphicx}
\usepackage{amsmath}
\usepackage{amssymb}
\usepackage{breqn}
\usepackage{cite}

\usepackage{enumitem}
\usepackage[percent]{overpic}

\usepackage[pagebackref=true,breaklinks=true,letterpaper=true,colorlinks,bookmarks=false]{hyperref}

\cvprfinalcopy 

\def\cvprPaperID{3573} 

\def\figvspace{{\vspace{-6mm}}}

\newcommand{\Sim}{\mathord{\sim}}
\def\eqnvspace{{\vspace{-3mm}}}
\newcommand{\Paragraph}[1]{\vspace{1.25mm} \noindent \textbf{#1} \hspace{0mm}}

\newcommand{\xdownarrow}[1]{%
  {\left\downarrow\vbox to #1{}\right.\kern-\nulldelimiterspace}
}

\ifcvprfinal\fi
\begin{document}

\title{Pedestrian Detection with Autoregressive Network Phases}

\author{Garrick Brazil, Xiaoming Liu \\
Michigan State University, East Lansing, MI \\
{\tt\small \{brazilga, liuxm\}@msu.edu}
}

\maketitle

\begin{abstract}
We present an autoregressive pedestrian detection framework with cascaded phases designed to progressively improve precision.
The proposed framework utilizes a novel lightweight stackable decoder-encoder module which uses convolutional re-sampling layers to improve features while maintaining efficient memory and runtime cost. 
Unlike previous cascaded detection systems, our proposed framework is designed within a region proposal network and thus retains greater context of nearby detections compared to independently processed RoI systems.
We explicitly encourage increasing levels of precision by assigning strict labeling policies to each consecutive phase such that early phases develop features primarily focused on achieving high recall and later on accurate precision.
In consequence, the final feature maps form more peaky radial gradients emulating from the centroids of unique pedestrians.
Using our proposed autoregressive framework leads to new state-of-the-art performance on the reasonable and occlusion settings of the Caltech pedestrian dataset, and achieves competitive state-of-the-art performance on the KITTI dataset.

\end{abstract}

\input{sec_1.tex}
\input{sec_2.tex}

\input{sec_3.tex}

\input{sec_4.tex}
\input{sec_5.tex}

{\small
\bibliographystyle{ieee}
\bibliography{egbib}
}

\end{document}

%% file: sec_1.tex
\section{Introduction}

\begin{figure}[t]
\begin{center}
   \includegraphics[width=1\linewidth]{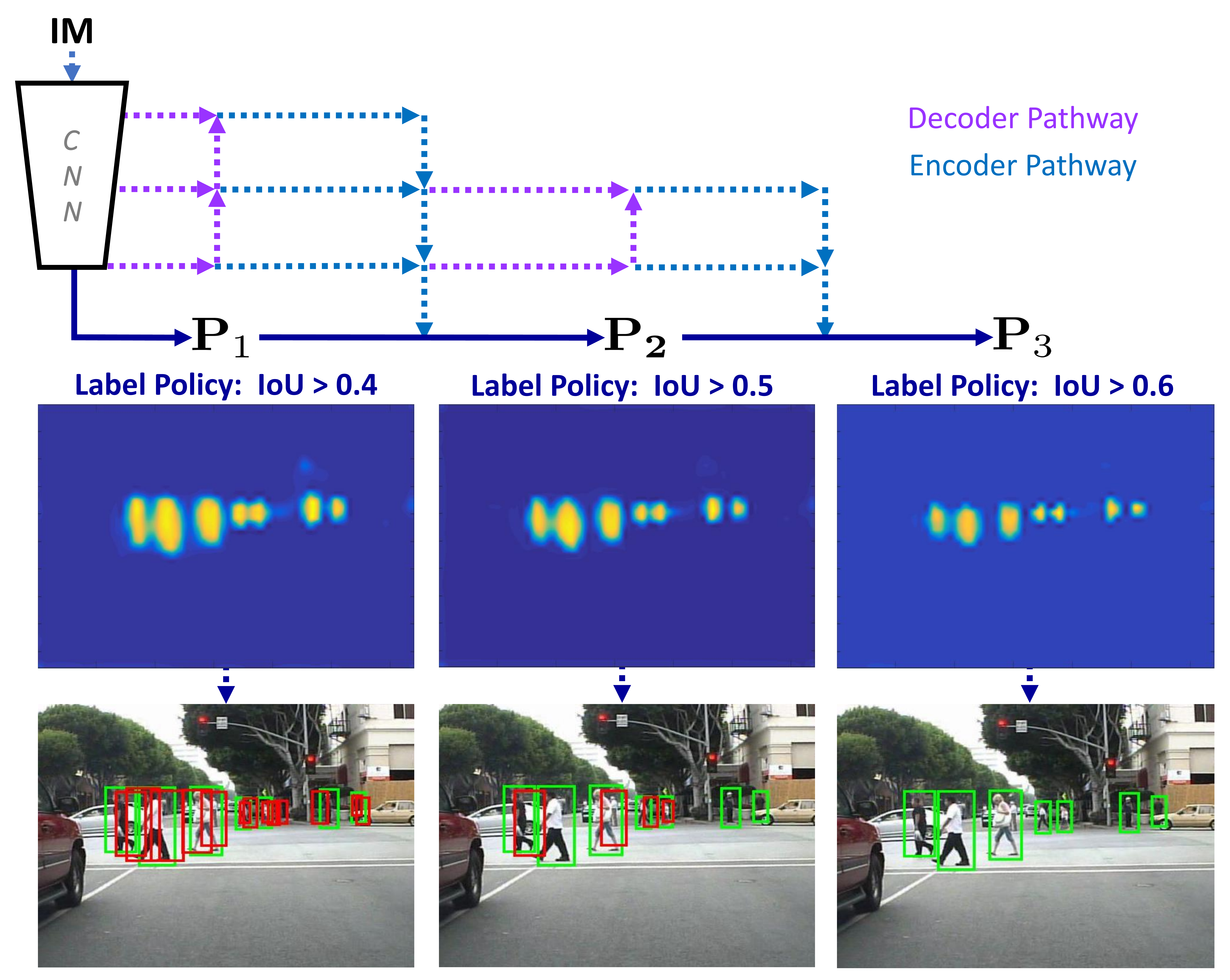}
      \caption{
Illustration of our proposed autoregressive framework with sample phase (P$_{1\to3}$) classification prediction maps and box visualizations under Caltech~\cite{dollar2009pedestrian} dataset. 
Our method iteratively re-scores predictions under incrementally more precise label policies, using a series of de-encoder modules comprised of decoder and encoder pathways.
Notice a heavy reduction in false positives (red) as phases progress, while true positives (green) are retained.
}
\label{fig:intro_fig}
\end{center}\vspace{-6mm}
\end{figure}

Detecting pedestrians in urban scenes remains to be a challenge in computer vision despite recent rapid advances~\cite{brazil2017illuminating, kimimproving, liu2018learning, song2018small, ren2017accurate, wang2017repulsion, zhou2018bi, zhang2018occlusion, zhang2018occluded}. 
The use of ensemble~\cite{du2016fused, szegedy2015going, Zagoruyko2016Multipath} and recurrent~\cite{ren2017accurate, stewart2016end} networks has been successful in top-performing approaches of pedestrian / object detection.
Recurrent networks refine upon their own features while ensemble networks gather features through separate deep classifiers.
Both techniques offer a way to obtain stronger and more robust features, thus better detection.

However, the characteristics of ensemble and recurrent networks are distinct.
Ensemble networks assume that separate networks will learn diversified features which when combined will become more robust.
In contrast, recurrent networks inherit previous features as input while further sharing weights between successive networks.
Hence, recurrent networks are more capable of refining than diversifying. 
Intuitively, we expect that both feature \textit{diversification} and \textit{refinement} are important components to pair together.

Therefore, we explore how to approximate an ensemble of networks using a stackable lightweight decoder-encoder module and incorporating an \textit{autoregressive}\footnote{We adopt naming distinction of \textit{autoregressive} (vs.~recurrent) as a network conditioned on previous predictions without the constraint of repeated shared weights, inspired by terminology in WaveNet~\cite{van2016wavenet} which uses casual convolution instead of conventional recurrence.} flow to connect them, as illustrated in Fig.~\ref{fig:intro_fig}.
We formulate our framework as a series of phases where each is a function of the previous phase feature maps and classification predictions.
Our decoder-encoder module is made of bottom-up and top-down pathways similar to~\cite{kong2018deep, lin2017feature, liu2018path, newell2016stacked}.
However, rather than using bilinear or nearest neighbor re-sampling followed by conventional convolution, we propose memory-efficient convolutional re-sampling layers to generate features and re-sample \textit{simultaneously} in a single step. 

In essence, our approach aims to take the best world of both the ensemble and recurrent approaches. 
For instance, since past predictions and features are re-used, our network is able to refine features when necessary.
Secondly, since our phases incorporate inner-lateral convolutions and do \textit{not} share weights, they are also capable to learn new and diversified features. 
Furthermore, we are able to design the network with an efficient overhead due to the added flexibility of using non-shared network weights for each phase and by using memory-efficient convolutional re-sampling layers. 
As a consequence, we are able to choose optimal channel settings with respect to efficiency and accuracy. 

To take full advantage of the autoregressive nature of our network, we further assign each phase a distinct labeling policy which iteratively becomes more \textit{strict} as phases progress.
In this way, we expect that the predictions of each consecutive phase will become less noisy and produce tighter and more clusterable prediction maps.
Under the observation that our proposed autoregressive region proposal network (RPN) obtains a high recall in the final phase, we also incorporate a simple hard suppression policy into training and testing of our second-stage R-CNN classifier. 
Such a policy dramatically narrows the subset of proposals processed in the second-stage pipeline ($\Sim 65\%$), and greatly alleviates the runtime efficiency accordingly. 





We evaluate our framework on the Caltech~\cite{dollar2009pedestrian} pedestrian detection dataset under challenging occlusion settings, using both the original and newly proposed~\cite{zhang2016far} annotations, and further on the KITTI~\cite{Geiger2012CVPR} benchmark.
We achieve state-of-the-art performance under each test setting and report a marginal overhead cost in runtime efficiency.

\noindent To summarize, our contributions are the following:
\begin{itemize}[noitemsep,topsep=1mm,label=$\bullet$]
\setlength\itemsep{1mm}
\item We propose a multi-phase autoregressive pedestrian detection system \textit{inside} a RPN, where each phase is trained using increasingly precise labeling policies.
\item We propose a lightweight decoder-encoder module to facilitate feature map refinement and message passing using convolutional re-sampling layers for memory-efficient feature pathways. 
\item  We achieve state-of-the-art performance on Caltech~\cite{dollar2009pedestrian} under various challenging settings, and competitive performance on KITTI~\cite{Geiger2012CVPR} pedestrian benchmark.
\end{itemize}

%% file: sec_2.tex
\section{Related Work}




\begin{figure}[t!]
\begin{center}
   \includegraphics[width=1\linewidth]{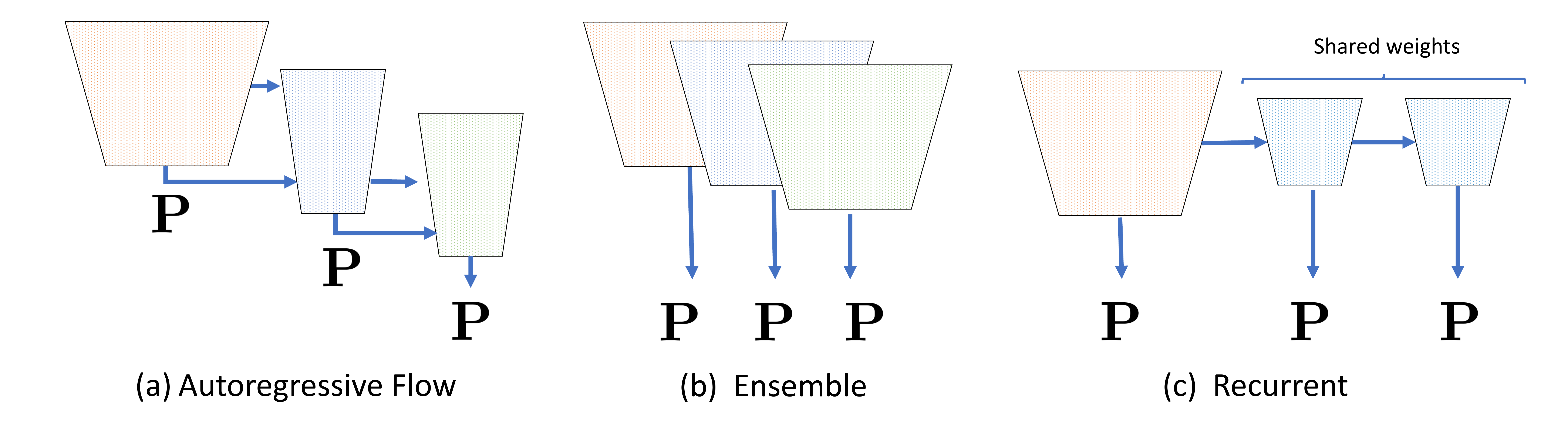}
   \caption{
Predictions of our autoregressive network (a) are directly conditioned on past feature maps as recurrent network (c) and do not share weights between phases as ensemble network (b). Unlike either, our network is further conditioned on past predictions.
}
\label{fig:compare}\figvspace
\end{center}\eqnvspace
\end{figure}

\begin{figure*}[t!]
\vspace{-2mm}
\begin{center}
   \includegraphics[width=0.90\linewidth]{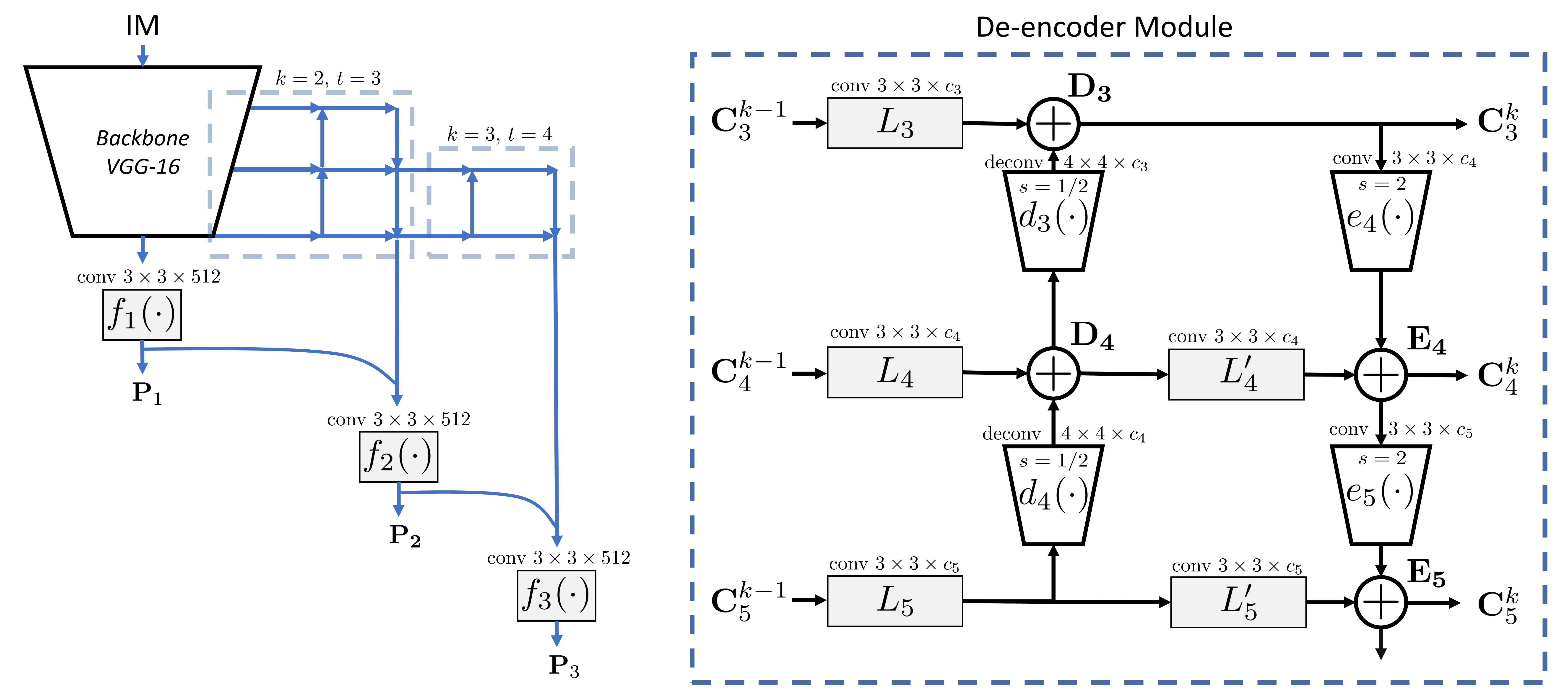}
   \caption{
Overview of our proposed AR-RPN framework (left) and detailed illustration of our de-encoder module (right). 
The de-encoder module consist of top-down and bottom-up pathways with inner-lateral convolution between pathways to produce diversified features, as well as convolutional re-sampling layers ($s$ denotes convolutional stride) $e_i$ and $d_i$ for memory-efficient feature generation. 
We further condition predictions on the previous phase predictions through concatenation within $f_k(\cdot)$.
 }
\label{fig:overview}
\end{center}\vspace{-5.5mm}
\end{figure*}

\Paragraph{Ensemble Networks:}
Recent top-performing methods~\cite{du2016fused, szegedy2015going, Zagoruyko2016Multipath} on detection have employed ensemble-based techniques where predictions from multiple deep convolutional neural networks (CNNs) are fused.
For instance, \cite{du2016fused} propose a soft-weighting scheme using an ensemble of independent detectors, which demonstrate high accuracy with fused scores.
However, one drawback is having multiple CNNs in memory and processing each in parallel.
Thus, both the scalability as networks become larger and usability in memory-constrained systems are lessened. 
Further,~\cite{brazil2017illuminating} form a small ensemble by fusing RPN scores with the scores of a R-CNN detector and demonstrates improved performance.
Compared to these methods, our \textit{single} RPN functions as an ensemble of inter-connected small networks, which can improve the precision  without critically obstructing runtime or memory efficiency.  

\Paragraph{Cascaded Networks:} A similar line to ensemble networks take form of cascaded detection systems~\cite{cai18cascadercnn,ouyang2017chained,qin2016joint}, which build on a series of R-CNN detectors and function on cropped region-of-interests (RoIs) generated by a static proposal network.
In contrast, our work focuses as a \textit{fully convolutional} cascade inside a {proposal network}. 
Therefore, our network is more equipped to utilize contextual cues of surrounding detections to inform suppression of duplicate detections, whereas cropped RoIs are processed \textit{independently} of other proposals. 
Liu \textit{et al.}~\cite{liu2018learning} propose supervision using incremental labeling policies similar to our approach. 
However, rather than making immediate predictions based only on previous predictions, we develop \textit{new} features through our decoder-encoder pathway.

\Paragraph{Recurrent Networks:}
Recurrent networks are a powerful technique in many challenging procedural~\cite{graves2009offline,mccann2017learned} and temporal~\cite{carreira2017quo,shi2017end,song2017end} computer vision problems.
Recently, it has been further demonstrated in urban object detection~\cite{ren2017accurate} and person head detection~\cite{stewart2016end}.
For instance, \cite{stewart2016end} uses recurrent LSTM to iteratively detect a single person at a time until reaching an end condition, thus side-stepping the need to perform non-maximum suppression (NMS) in post.
In contrast,~\cite{ren2017accurate} proposes a rolling recurrent convolution (RRC) model which refines feature maps and produces new detections at each step.
From this respect, our proposed method is similar to RRC, but with two critical differences. 
Firstly, the networks of our phases are not shared. 
This enables us to learn specialized (ensemble-like) features in each phase and gives more freedom in network design of a phase, which may aid runtime efficiency when using conservative designs. 
Secondly, we base each phase conditioned on previous feature maps \textit{and} predictions, which form a more potent autoregressive foundation.
We show a high-level comparison of our autoregressive network, ensemble networks, and recurrent networks in Fig.~\ref{fig:compare}.

\Paragraph{Encoder-Decoder Networks:}
Many recent works~\cite{kong2018deep, lin2017feature, liu2018path, ronneberger2015u} have explored multi-strided feature maps re-use within computer vision. 
Each variant of architectures utilize a series of convolution, feature aggregation (concat, residual), and up-sampling / pooling layers in order to form an encoder-decoder structure.
Similar to the network structure in~\cite{newell2016stacked} for human pose estimation, we incorporate stackable top-down \textit{and} bottom-up pathways. 
However, in contrast to prior work, we design our de-encoder module without explicitly using bilinear or nearest neighbor re-sampling.
Instead, we uniquely blend the feature generation and re-sampling into a single convolution layer using a fractional stride ($\uparrow$) or strided convolution ($\downarrow$), making the travel nodes in our streams as \textit{compact} as possible.
We show in ablation that a single convolutional re-sampling layer consumes low memory and performs better compared to the conventional two-step techniques previously used.

%% file: sec_3.tex
\section{Autoregressive Detector}

Our proposed framework is made up of two stages: an autoregressive RPN hence referred to as AR-RPN, and a second-stage R-CNN detector each founded on~\cite{zhang2016faster, ren2015faster}.
We collectively refer to both stages as AR-Ped.  
As shown in Fig.~\ref{fig:overview}, AR-RPN consists of multiple phases, where each predicts classification scores and passes these predictions \textit{and} their features into the next phase. 
Each phase is connected to the last through a bottom-up and top-down pathways, which form a lightweight decoder-encoder module.
This module is stackable onto the backbone RPN and onto itself repeating.
We supervise each phase to jointly learn increasingly more precise predictions by imposing a stricter labeling policy to consecutive phases, thereby producing more peaky and clusterable classifications in the final phase.
We apply the box transformations, NMS, and a hard suppression policy to the final predictions for which the remaining subset are used to train a specialized R-CNN detector.

\subsection{De-Encoder Module} 

To perform autoregressive detection in a {\it single} model, we design a stackable decoder-encoder module, termed \textit{de-encoder}, where its top-down pathway leverages past feature maps and its bottom-up pathway encodes stronger semantics.
Following~\cite{lin2017feature}, we give each pathway the ability to learn from feature maps at multiple depths of the backbone network. 
Importantly, our design encourages the highest level features to remain at the lowest resolution where object detection functions most efficiently.
Intuitively, the de-encoder enables the network to look back at previous features \textit{and} learn more advanced features during re-encoding.

Let us recall that typical network architectures, e.g., VGG-16~\cite{simonyan2014very} and ResNet-50~\cite{he2016deep}, function from low to high stride levels using a series of convolution and pooling layers. 
We denote the set of strides of a backbone network as $\mathbf{S}$, where $2^{i-1}$ is the down-sampling factor of the $i$th stride level preceding a pooling operation. 
In pedestrian detection, it is common to have $n=5$ unique stride levels such that $\mathbf{S} = \{1, 2, 4, 8, 16\}$. 
The hyperparameters of the de-encoder module include a designated target stride $t$ and channel width $c_i$ specific to each stride, which respectively control how far up in resolution the phase should de-encode and how many channels at each stride should be learned. 

The primary goal of the de-encoder module is to produce finer features at each level from the target stride $t$ to the final stride $n$ of the network. 
Denoting $\mathbf{C}^k = \{\mathbf{C}^k_t, \dots, \mathbf{C}^k_n\}$ as the refined features of the $k$th phase at each stride, $g _k(\cdot)$ as the set of convolutional and ReLU layers, $\Phi_k$ the respective weights, and~$t_k$ the target stride of feature maps to de-encode and refine, 
the autoregressive nature of the feature generation can be expressed as: 
\begin{equation}
\eqnvspace
\mathbf{C}^k = g_k (\mathbf{C}^{k-1}\ |\ \Phi_k ,\ t_k).  
\eqnvspace
\end{equation}
Hence, each phase of the network takes as input the previous phase feature maps and produces more advanced features. 
Initial features $\mathbf{C}^1$ are given from top-most layers at corresponding strides from the backbone (e.g., in VGG-16 $\mathbf{C}^1_4=$~conv4\_3, $\mathbf{C}^1_5=$~conv5\_3, and so forth).  



\Paragraph{Top-down pathway:} 
We design our top-down decoder for phase $k$ by attaching a convolutional layer with BN~\cite{ioffe2015batch} to $\{\mathbf{C}^{k-1}_{t\dots n}\}$ feature maps, which produce inner-lateral convolutions $\mathbf{L}_{i}$ with corresponding channel widths $c_i$.
Rather than using a two-step process comprised of a bilinear / nearest neighbor up-sampling followed by convolution as done in prior work, we denote $d_i(\cdot)$ as a convolutional up-sampling layer which \textit{simultaneously} performs $2\times$ up-sampling and feature reduction into channel width $c_{i}$ using fractionally strided convolution.
The combined operation is more efficient in both memory and runtime.
Starting with the highest feature stride $n$, we use $d_i(\cdot)$ to iteratively decode features, which are then fused with the lateral features at the decoded stride $\mathbf{L}_{i}$ through element-wise addition, denoted:  
\begin{equation}
\mathbf{D}_i = d_i(\mathbf{D}_{i+1})+ \mathbf{L}_{i}.
\label{eqn:top-down}
\end{equation}
We begin with the base case of $\mathbf{D}_n=\mathbf{L}_n$, and repeat this procedure until the target stride feature map $\mathbf{D}_t$ is reached.
In theory, the top-down pathway enables high-level semantics to be passed down through the decoded term $d_i(\mathbf{D}_{i+1})$ and low-level features to be re-examined using $\mathbf{L}_{i}$.  

\Paragraph{Bottom-up pathway:} 
We design the bottom-up encoder in the opposite manner as the decoder.
We first attach a convolutional layer with BN to each $\{\mathbf{D}^{k-1}_{t+1\dots n}\}$ which each produce \textit{new} laternal features $\mathbf{L}^\prime_i$ with $c_i$ channels.
Similar to the decoder pathway, we denote $e_i(\cdot)$ as a single convolutional down-sampling layer which \textit{simultaneously} performs $2\times$ down-sampling and feature expansion into channel width $c_{i}$ using strided convolution, rather than conventional two-step process used in previous work.  
We use $e_i(\cdot)$ to iteratively encode the features at each stride, which are then fused with the lateral features of the encoded stride  $\mathbf{L}^\prime_{i}$ via element-wise addition, denoted as:   
\begin{equation}
\eqnvspace
\mathbf{E}_i = e_i(\mathbf{E}_{i-1})+ \mathbf{L}^\prime_{i}.
\eqnvspace
\label{eqn:bottom-up}
\end{equation}
As the name suggests, the bottom-up encoder starts with the lowest stride $t$ and repeats until the $n$th stride is reached, such that lateral features at $t$ is $\mathbf{E}_t = \mathbf{D}_t$.
The bottom-up pathway enables the network to encode low-level features from the lowest stride through the $e_i(\mathbf{E}_{i-1})$ term and for higher-level features to be re-examined using $\mathbf{L}^\prime_{i}$.

\subsection{Autoregressive RPN}
\label{ss:autorpn}

We utilize the standard RPN head and multi-task loss proposed in~\cite{girshick2015fast} following the practices in~\cite{zhang2016faster}.
We predefine a set of anchor shapes which act as hyperparameters describing the target pedestrian scales. 
The RPN head is comprised of a proposal feature extraction (PFE) layer connected to two sibling layers which respectively predict anchor classification (cls) and bounding box regression (bbox) output maps, hence forming a multi-task learning problem. 

\Paragraph{Multi-phase Network:}
Our RPN is comprised of a total of $N_k = 3$ phases.
The first phase is simply the backbone network starting with the modified VGG-16~\cite{simonyan2014very} that has strides of $\mathbf{S} = \{1, 2, 4, 8, 16\}$.
The second phase is a de-encoder module which has a target stride $t = 3$ and channel widths of $c_3 = 128, c_4=256, c_5=512$.
The final phase is another stack of the de-encoder module following the same channel settings but uses a memory conservative lower target stride of $t=4$.
The spatial resolution at $i$th stride can be denoted as $w_i \times h_i = \frac{W}{2^{i-1}} \times \frac{H}{2^{i-1}}$, where $W \times H$ is the input image resolution.  
Thus, the final proposal network architecture forms a stair-like  shape as in Fig.~\ref{fig:overview}.  

\Paragraph{Autoregressive Flow:}
To enable the autoregressive flow between phases, we place a PFE layer and classification layer at the end of each phase encoder. 
For all phases except the first, we \textit{concatenate} the previous phase predictions into the input features for the corresponding phase PFE layer. 
In doing so, each phase is able to start with strong compact features by directly utilizing its previous phase predictions. 
Further, the PFE layer of the final phase $N_k$ produces the bounding box regression output map, since these features are the most precise and \textit{peaky} within the network.  

Formally, we denote functions $f_k(\cdot)$ and $p_k(\cdot)$ as the $k$th phase PFE layers and classification layers respectively.
We build $f(\cdot)$ as a convolutional layer with $3\times 3$ kernel and $512$ output channels followed by a ReLU layer, while $p(\cdot)$  a convolutional layer with $1\times 1$ kernel and outputs channels $2\times$ the number of anchors ($A$).  
Thus, $p_k(\cdot)$ forms an autoregressive function of previous phase predictions with an output dimension of $w_5 \times h_5 \times 2A$, via:   
\begin{eqnarray}
\eqnvspace
\mathbf{P}_k = p_k(f_k(\mathbf{P}_{k-1}\ \|\ \mathbf{C}^k_n)),
\eqnvspace
\end{eqnarray}
where $\mathbf{P}_{k-1}$ is the classification feature map of the previous phase, aka, past predictions, $\|$ is the concatenation operator, and $\mathbf{C}^k_n$ is the last encoded feature map of the $k$th phase.  
As defined, the PFE $f_k(\cdot)$ and classification layer $p_k(\cdot)$ are conditioned autoregressively on past predictions which logically act as compact but powerful semantic features.
In this way, each phase is more free to learn new features $\textbf{C}^k_n$ to directly {\it complement} the past predictions. 
In essence, the autoregressive flow can be seen as running memory of the most compact and strong features within the network.  

\Paragraph{Classification Task:}
Each classification layer which proceeds a PFE layer is formulated as proposed in~\cite{ren2015faster} following experimental settings of~\cite{brazil2017illuminating}. 
Formally, given a PFE layer with dimensions~$w\times h$, the designated classification layer predicts a score for \textit{every} spatial location of the image $(x,y) \in \mathbb{R}^{w \times h}$ against every predefined anchor shape $a \in \mathbf{A}$, and every target class. 
Every spatial location of the prediction map is therefore treated as a distinct box with its own corresponding classification score.
To produce labels for each box, a labeling policy is adopted using a hyperparameter $h$ that controls the box criteria of Intersection over Union (IoU) with ground truths in order to be considered foreground.
After every box is assigned a label according to the labeling policy, each classification layer is supervised using multinomial cross-entropy logistic loss as in~\cite{girshick2015fast}.

\Paragraph{Localization Task:}
The localization task is formed using the same set of anchor boxes described in the classification task. 
The localization task aims to perform bounding box regression that predicts a bounding box transformation for each foreground box towards the nearest pedestrian.
A proposal box is considered \textit{nearby} a pedestrian ground truth if there is at least $h$ intersection over union between the two boxes. 
The box transformation is defined by $4$ variables consisting of translation $(t_x, t_y)$ and scale factors $(t_w, t_h)$ such that when applied will transform the source box into the target ground truth.
We train the bounding box regression values using Smooth $L_1$ loss~\cite{girshick2015fast}.  

\Paragraph{Incremental Supervision:}
In order to better leverage the autoregressive and de-encoder properties of AR-RPN, we choose to assign \textit{different} classification labeling policies onto each consecutive phase. 
We emphasize that the de-encoder modules enable the network to adapt and become a stronger classifier, which can be exploited to produce more accurate and tighter classification clusters when supervised with incrementally stricter labeling policies.

Let us briefly discuss the trade-offs regarding different labeling policies.
Consider using a labeling policy of $h = 1$, which is approximately equivalent to requiring the network output a \textit{single} box for each pedestrian and thus the imbalance of classes may be difficult. 
In contrast, as a labeling policy becomes more lenient at $h = 0.5$, the classification becomes more balanced but produces many false positives as duplicate detections. 
In theory, bounding box regression will reduce the impact of double detections by transforming boxes into clusters which can be suppressed by NMS.
Ideally, a network has either high-performing bounding box regression and/or tight clusterable classification maps, since both enable NMS to cluster duplicate detections.
Therefore, rather than using a single discrete labeling policy of $h=0.5$, we assign lenient-strict policies $h_1=0.4,~h_2=0.5,~h_3=0.6$, to each phase classification layer respectively. 
In contrast to~\cite{liu2018learning}, we enforce incremental supervision between de-encoder modules rather than being applied immediately in quick succession. 
In consequence, our classification score maps are supervised to gradually become more peaky and clusterable. 

\Paragraph{Loss Formulation:} 
In addition to the classification and bounding box regression losses, we further add auxiliary losses in the form of weak semantic segmentation as in~\cite{brazil2017illuminating}. 
Specifically, during training we add a binary semantic segmentation layer to each stride of the first top-down pathway to act as an auxiliary loss and accelerate training.
We formally define the joint loss terms incorporating phase classification softmax loss $L_{cls}$, final phase localization Smooth $L_1$ loss $L_{bbox}$, and each softmax auxiliary loss $L_{seg}$ as: 
\begin{equation}
\eqnvspace
 L = \sum_{k=1}^{N_k} \lambda_{k} L_{cls}  + \lambda_{b} L_{bbox}  + \lambda_{s} \sum_{i=3}^5 L_{seg},
\eqnvspace
\end{equation}
where $k$ corresponds to phases $1\to N_k$ of the full network, and $i$ represents stride for each auxiliary segmentation layer of the backbone network.  
We use Caffe~\cite{jia2014caffe} with SGD following the settings in~\cite{zhang2016faster} in our training. 
We set $\lambda_1 = \lambda_2 = 0.1, \lambda_3 = 1$, $\lambda_b = 5$,  and $\lambda_s = 1$.

\begin{table*}[t!]
\begin{center}
\small
\setlength\tabcolsep{5.25pt}
  \resizebox{\textwidth}{!}{  
\begin{tabular}{l|c c c c  | c c | c  | c c c}
\hline
& \multicolumn{4}{c}{Caltech Reasonable} & \multicolumn{2}{c}{Caltech Occlusion} &  & \multicolumn{3}{c}{KITTI} \\ 
\hline
& $MR^O_{-2}$ & $MR^O_{-4}$ & $MR^N_{-2}$ & $MR^N_{-4}$ & Partial$^O$ & Heavy$^O$ & RT (ms)& Easy & Mod. & Hard \\ 
\hline\hline
MS-CNN~\cite{cai2016unified} & $9.95$ & $22.45$ & $8.08$ & $17.42$  & $19.24$ & $59.94$& $64$ & $\it{83.92}$ & $\it{73.70}$ & $\bf{68.31}$ \\
RRC~\cite{ren2017accurate} & $-$ & $-$ & $-$ & $-$ & $-$ & $-$  & $75$ & $-$ & $\bf{75.33}$ & $-$ \\
RPN+BF~\cite{zhang2016faster} & $9.58$ & $18.60$ & $7.28$  & $16.76$ & $24.23$ & $74.36$ & $88$  & $75.58$ & $61.29$ & $56.08$\\ 
F-DNN~\cite{du2016fused} & $8.65$ & $19.92$ & $6.89$  & $\it{14.75}$ & $15.41$ & $\it{55.13}$  & $-$ & $-$ & $-$ & $-$ \\ 
TLL(MRF)+LSTM~\cite{song2018small} & ${7.40}$ & $-$ & $-$  & $-$ & $-$& $-$& $-$& $-$& $-$& $-$ \\
ALFNet~\cite{liu2018learning} & $-$ & $-$ & $6.10$  & $-$ & ${-}$& $-$& $-$ & $-$ & $-$ & $-$\\
SDS-RCNN~\cite{brazil2017illuminating} & $\it{7.36}$ & $\it{17.82}$ & $6.44$  & $15.76$ & ${14.86}$& $58.55$& $95$ & $-$ & $63.05$ & $-$\\
\hline
RepulsionLoss~\cite{wang2017repulsion} & $-$ & $-$ & $\it{5.00}$ & $-$ & $-$& $-$& $-$& $-$& $-$& $-$ \\
FRCNN+ATT-vbb~\cite{zhang2018occluded} & $10.33$ & $-$ & $-$ & $-$ & $-$ & $45.18$ & $-$ & $-$ & $-$ & $-$\\
PDOE+RPN~\cite{zhou2018bi} & $7.60$ & $-$ & $-$ & $-$& $\it{13.30}$ & ${44.40}$& $-$  & $-$& $-$& $-$\\ 
GDFL~\cite{lin2018graininess} & $7.85$ & $19.86$ & $-$  & $-$ & $16.74$ & $\it{43.18}$  & $-$ & $\bf{84.61}$ & $68.62$ & $66.86$ \\ 
DSSD~\cite{fu2017dssd}+Grid~\cite{kimimproving} & $10.85$ & $18.20$ & $-$& $-$ & ${24.28}$ & $\bf{42.42}$& $-$& $-$& $-$& $-$ \\ 
\hline\hline
AR-RPN (ours) & $8.01$ & $21.62$ & $5.78$ & $15.86$ & $16.30$ & $58.06$  & $86$ & $-$ & $-$ & $-$ \\
AR-Ped (ours) & $\bf{6.45}$ & $\bf{15.54}$ & $\bf{4.36}$ & $\bf{11.39}$ & $\bf{11.93}$ & ${48.80}$& $91$ & $83.66$ & ${73.44}$ & $\it{68.12}$  \\
\hline
\end{tabular}
}
\end{center}
\caption{
Comprehensive comparison of our frameworks and the state-of-the-art on the Caltech and KITTI benchmarks, in both accuracy and runtime (RT).
We show the Caltech miss rates at multiple challenging settings, with both the original ($O$) and new ($N$) annotations, and at occlusion settings with the original annotations and FPPI range $MR^O_{-2}$.
Further, we evaluate the KITTI pedestrian class under easy, moderate, and hard settings, with mean Average Precision (mAP)~\cite{Geiger2012CVPR}. {\bf Boldface}/{\it italic} indicate the best/second best performance.
}
\vspace{-2mm}
\label{eval}
\end{table*}

\subsection{R-CNN Detector}

Most pedestrian detection frameworks are derivatives of Faster R-CNN~\cite{ren2015faster}, and hence incorporate a second-stage scale-invariant region classifier termed as R-CNN. 
Following~\cite{brazil2017illuminating}, we utilize a modified VGG-16 as a R-CNN that functions on cropped RGB regions proposed by AR-RPN, utilizes a strict labeling policy, and fuses its scores with the RPN. 
However, unlike past methods we impose a simple hard suppression policy that suppresses all box proposals with a score less than a hyperparameter $z$.
This has two advantages.
Firstly, it greatly improves runtime since only a subset of proposals need to be processed.
Secondly, by \textit{focusing}  on only the hard samples leftover from the RPN, the R-CNN learns specialized classification similar to the motivation of the AR-RPN. 

\Paragraph{Loss Formulation:}
As in the AR-RPN, we also use softmax loss to train the R-CNN.
We use a strict labeling policy requiring $h\geq 0.7$ IoU for foreground, a weak segmentation auxiliary loss $L_{seg}$, and height sensitive weighting scheme $w$ as detailed in~\cite{brazil2017illuminating}.
We set $z=0.005$ to impose a score suppression of the RPN proposals and eliminate confident background proposals from being re-processed. 
In practice, the suppression dramatically reduces the search space for both efficiency and accuracy while critically keeping recall \textit{unaffected}.
Thus, we denote the R-CNN loss as: 
\begin{equation}
\eqnvspace
 L = \sum_j w_j L_{cls} (c_j, \hat{c_j}) + L_{seg}, \hspace{0.5cm} \text{if } c_j\geq z,
\eqnvspace
\end{equation}
where $j$ corresponds to each proposal of AR-RPN, $c$ is the classification result of the R-CNN, and $\hat{c}$ is the class label. 
We use Caffe to train the R-CNN following settings of~\cite{brazil2017illuminating}.



%% file: sec_4.tex
\section{Experiments} 

We evaluate our proposed AR-Ped framework on two challenging datasets: Caltech~\cite{dollar2009pedestrian,dollar2012pedestrian} and KITTI~\cite{Geiger2012CVPR}. 
We perform experiments ablating our approach from the perspective of design choices and hyperparameters.
We further examine the qualitative changes and analyze the quantitative {peakiness} in detections across phases. 

\subsection{Caltech}

The Caltech~\cite{dollar2009pedestrian,dollar2012pedestrian} dataset is a widely used benchmark on pedestrian detection that  contains $10$ hours of video taken from an urban driving environment with  $\sim$$350{,}000$ bounding box annotations and $2{,}300$ unique pedestrians.
We use the Caltech$10\times$ for training and the Caltech reasonable setting~\cite{dollar2012pedestrian} for testing, unless otherwise specified. 
The evaluation uses a miss rate (MR) metric averaged over a false positive per image (FPPI) range of $[10^{-2}, 10^0]$ and also a more challenging metric over the range $[10^{-4}, 10^0]$, respectfully referred to as $\textit{MR}_{-2}$ and $\textit{MR}_{-4}$. 
Recently, new annotations are released~\cite{zhang2016far} to correct the official annotations in terms of consistency and box alignment.
For completeness, we evaluate on both the original and the new annotations, denoted respectively as $\textit{MR}^{O}$ and $\textit{MR}^{N}$.

We compare our work to the state-of-the-art pedestrian detection methods of Caltech with respect to the core experimental configurations of using each combination of original/FPPI setting, and partial/heavily occlusion within the original annotation space as defined in~\cite{dollar2009pedestrian}.
We limit our comparison to the top-$2$ methods of any sub-category trained using Caltech$10\times$ dataset since these comprise the most highly competitive methods. 
We also emphasize that we are among the few methods to \textit{comprehensively} evaluate and report each setting and insist to open-source our code to the community upon release.

Our method advances the state-of-the-art on all but one evaluation setting, as detailed in Table~\ref{eval}.
Under the most common benchmark reasonable setting, we achieve a miss rate of $6.45\%$ ($\downarrow 0.91$) and $4.36\%$ ($\downarrow 0.64$) on the official annotations $\textit{MR}^{O}_{-2}$ and new annotation $\textit{MR}^{N}_{-2}$ respectively.
Further, our approach has increased robustness to partial occlusion ($\downarrow 1.37\%$ miss rate).
Compared to methods which do not explicitly address occlusion~\cite{cai2016unified, brazil2017illuminating, du2016fused, zhang2016faster}, our method also improves w.r.t heavy occlusion ($\downarrow 6.33\%$ miss rate). 
Yet, our method underperforms on heavy occlusion compared to work specially designed to target occlusion problem~\cite{kimimproving, lin2018graininess, wang2017repulsion, zhou2018bi}, which is orthogonal to our work. 

We further produce a runtime analysis for state-of-the-art works with public code using the {\it same} controlled machine with NVIDIA $1080$ Ti GPU, as summarized in Table~\ref{eval}. 
Our method retains a competitive runtime efficiency due to the light overhead design of our de-encoder module while still improving accuracy in all but one setting. 

\subsection{KITTI}
KITTI is a popular urban object detection dataset which offers annotations for cars, pedestrians and cyclists.
We use the official training set of $7{,}481$ images and evaluate on the standard $7{,}518$ test images.
We adopt the settings and core training code of~\cite{cai2016unified} in order to initialize good starting hyperparameters. 
However, due to GPU memory constraints we set the input image scale to $576$ height resolution and achieve competitive performance on the pedestrian class, as reported in Table~\ref{eval}. 
As described in~\cite{brazil2017illuminating}, high performing pedestrian detectors~\cite{brazil2017illuminating,li2015scale,zhang2016faster} on Caltech and KITTI do not usually have high correlation.
We emphasize that \textit{our AR-Ped is among the first to report high performance for both datasets}, which suggests the generalization of our model to pedestrian detection rather than a specific dataset.

\subsection{Ablations}

All ablation experiments use our AR-RPN and the Caltech test set under the reasonable $MR^O_{-2}$ FPPI setting, as this is the most widely tested setting on Caltech. 

\Paragraph{What are optimal de-encoder settings?}
In order to analyze the de-encoder module, we ablate its parameters in each phase concerning channel widths at each feature stride and target strides to de-encode.
Our primary method of AR-RPN uses what we refer to as medium channel width settings of $\mathbf{c}_{M} = \{128,~256,~512\}$.
We further denote small and large channel settings such that $\mathbf{c}_{S} = \{64,~128,~256\}$ and $\mathbf{c}_{L} = \{256,~512,~512\}$, then train our AR-RPN with other settings kept consistent. 
Surprisingly, the small and large channel widths function similarly but neither as well as the medium, which roughly follows the rules-of-thumb channel settings outlined in VGG-16~\cite{simonyan2014very}. 
For instance, the $\mathbf{c}_{L}$ and $\mathbf{c}_{S}$ achieve $8.33\%$~($\uparrow 0.32\%$) and $8.62\%$~($\uparrow 0.61\%$) miss rate, as detailed in Table~\ref{tab:phase_ablation}.
This suggests a difficulty when over or under expanding channels compared to the $c$ width of source feature maps in $\mathbf{C}^1$.

We further analyze the runtime complexity of the de-encoder modules under each proposed setting in Table~\ref{tab:phase_ablation}.
Overall, we observe that channel width settings have a large effect on both multiply-accumulate (MAC) and runtime efficiencies of the AR-RPN.
Specifically, channel width settings of $\mathbf{c}_{S}$, $\mathbf{c}_{M}$, and $\mathbf{c}_{L}$ respectively slow down by $8\%$, $26\%$, and $69\%$ compared to $N_k = 1$ baseline.

\begin{table} [t]
\begin{center}
\setlength\tabcolsep{3.5pt}
\begin{tabular}{c c c c c c}
\hline
$N_k$ & $\mathbf{c}$ size & {$MR^O_{-2}$} & MAC (G) & Runtime (ms)\\
\hline
$1$ & $M$ & $10.16$  & $217.9$& ${68}$ \\
$2$ & $M$ & $8.32$ & $429.3$& $80$   \\
$3$ & $S$ & ${8.62}$ & $255.3$&  $74$  \\
$3$ & $M$ & ${8.01}$ & $321.3$& $86$   \\
$3$ & $L$ & ${8.33}$ & $429.3$& $115$   \\
$4$ & $M$ & ${8.68}$ & $355.9$& $97$  \\

\hline
\end{tabular}
\vspace{1mm}
\caption{The performance with different parameters and numbers of phases under the Caltech reasonable $MR^O_{-2}$ setting.
We further detail the efficiency of each setting in terms of multiply-accumulate (MAC) and runtime on an NVIDIA $1080$ Ti.
}\label{tab:phase_ablation}
\end{center}\vspace{-7mm}
\end{table}

\Paragraph{What is the effect of convolutional re-sampling?} Unlike  previous decoder-encoder works~\cite{kong2018deep, lin2017feature, liu2018path, newell2016stacked, ronneberger2015u}, our module combines its re-sampling and feature generation into single convolutional re-sampling layers using either stride of $2$ or fractional $\frac{1}{2}$ strides. 
To better understand the importance of this combined operation, we split every convolutional re-sampling layer $e(\cdot)$ and $d(\cdot)$ into $2$ separate layers: a bilinear re-sampling layer and a convolution feature generation layer. 
We observe that this separation causes performance to degrade from $8.01\% \to 9.45\%$ miss rate. 
This degradation suggests that providing the network with more freedom in re-sampling, as opposed to fixing the kernels to bilinear (or nearest neighbor), is beneficial for detection.
Moreover, separating the operations into $2$-steps is naturally less efficient concerning memory usage and runtime.
Specifically, using the proposed convolutional re-sampling layers within AR-RPN consumes $41\%$ less GPU memory compared to using a $2$-step bilinear / convolution process and maintains a $16\%$ faster runtime speed at inference.

\begin{figure*}[t]
\vspace{-2mm}
\begin{center}
   \includegraphics[width=0.90\linewidth]{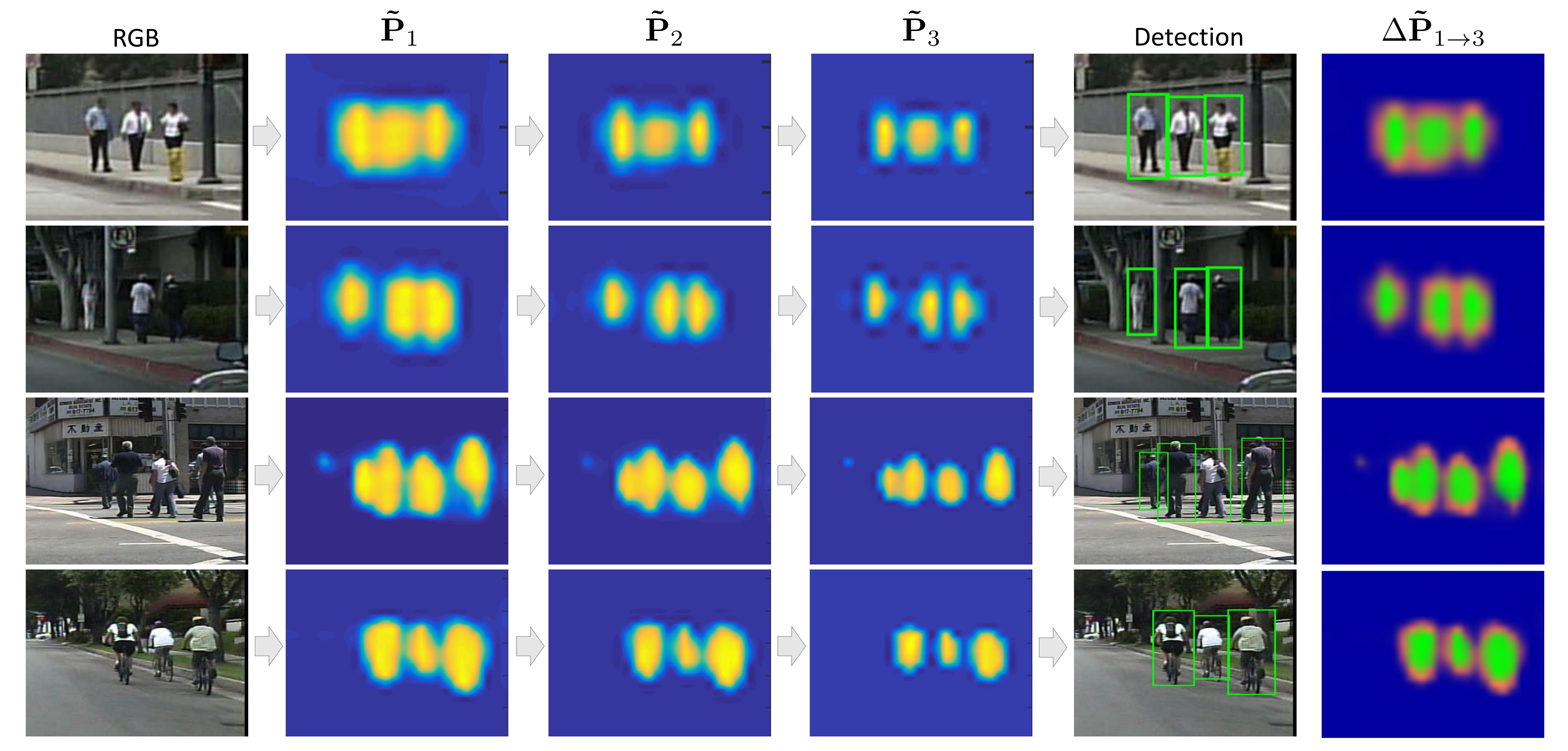}
\vspace{-1mm}
      \caption{
We visualize the prediction maps $\tilde{\mathbf{P}}_k$ of each phase by taking the maximum of foreground scores across all $A$ anchors at each spatial location, i.e., denoting $\mathbf{P}_k =\{\mathbf{P}_k^{bg}, \mathbf{P}_k^{fg}\}$, we define $\tilde{\mathbf{P}}_k = \max_{A} \mathbf{P}_k^{fg}$.
We use scaled blue~$\to$yellow colors to visualize $\tilde{\mathbf{P}}_k$, where yellowness indicates high detection confidence.
The detections of each phase become increasingly tighter and more adept to non-maximum suppression due to the incremental supervision for each phase (Sec.~\ref{ss:autorpn}).
We further analyze the prediction disagreements between phases $\Delta 1\to3$, shown in the right column, where green represents the agreement of the foreground and magenta the regions suppressed.
}
\label{phase_visualization}
\end{center}\vspace{-5mm}
\end{figure*}

\Paragraph{How many autoregressive phases to stack?}
The use of autoregressive phases is clearly a \textit{critical} component of our framework.
Therefore, to understand its impact we ablate our framework by varying the number of phases while keeping all other settings constant. 
We report the performance of each setting in Table~\ref{tab:phase_ablation}.
Unsurprisingly, as fewer phases are used the performance is steeply reduced.
For instance, recall that our $3$-stage method achieves $8.01\%$ miss rate. 
By removing a single phase, the miss rate increases by $\uparrow0.32\%$ while only gaining $6$ ms in runtime efficiency. 
When another phase is removed, an extreme degradation of $\uparrow2.15\%$ is observed.
Hence, the effect of additional phases seems to diminish with $N_k$ such that the first additional phase has the highest impact, as suggested by Fig.~\ref{phase_visualization}. 
We further add a $4th$ phase following the same trend in incremental labeling ($h_4 = 0.7$) and observe that the performance begins to worsen. We suspect using more dense anchor sampling may help train the very high IoU threshold. 

\begin{table} [t]
    \begin{center}
\setlength\tabcolsep{14pt}
\begin{tabular}{c|c}
\hline
Labeling Policy & {$MR^O_{-2}$} \\
\hline
no autoregressive & ${9.06}$ \\
~~~~strict~$\to$~lenient & $9.03$ \\
~~moderate~$\to$~moderate & $8.94$ \\
~~strict~$\to$~strict & $8.43$ \\
lenient~$\to$~strict & ${8.01}$ \\
\hline
\end{tabular}
\vspace{1mm}
\caption{The effects of labeling policies on the Caltech dataset under the reasonable $MR^O_{-2}$ setting.}\label{tab:labeling_policy}
\end{center} \vspace{-7mm}
\end{table}

\begin{figure}[t]
\begin{center}
   \includegraphics[width=0.975\linewidth]{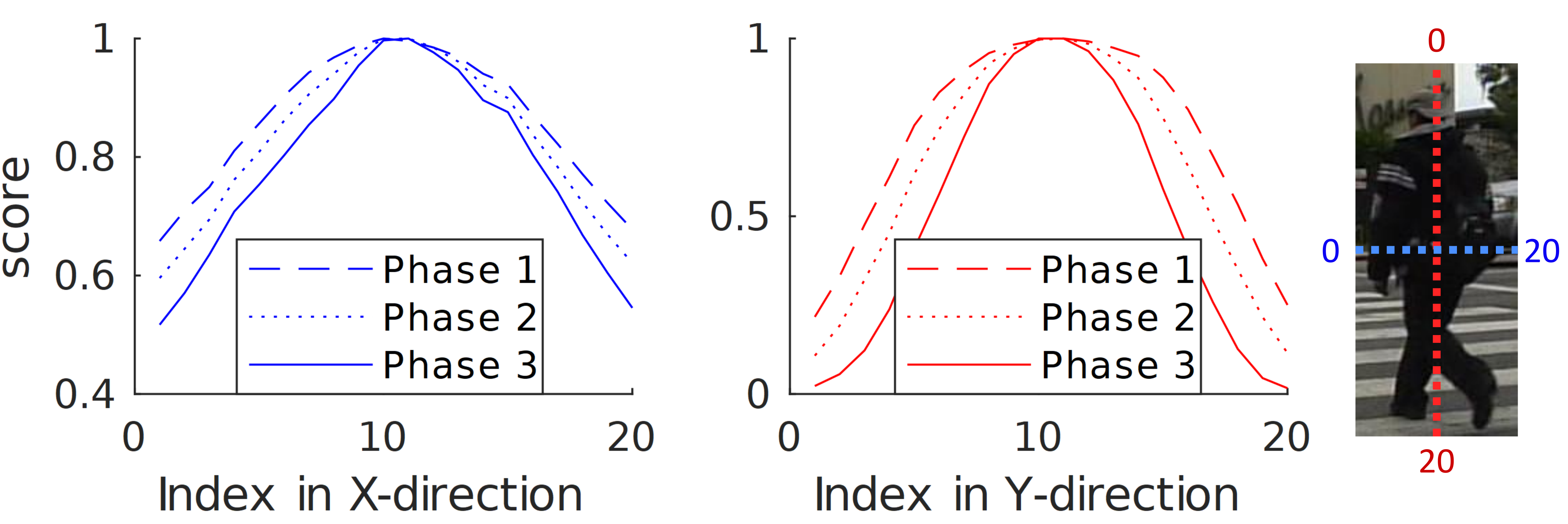}
      \caption{
We analyze the mean prediction score ($\tilde{\mathbf{P}}_k$) of $20$ uniformly sampled points along the center lines of X-direction (left) and Y-direction (right) averaged over \textbf{all} ground-truth pedestrians in Caltech test dataset, using bilinear interpolation when necessary. 
We note that successive phase scores form more \textit{peaky} inclines radiating from the center of the pedestrian. 
}
\label{p_scores}
\end{center}\vspace{-5mm}
\end{figure}

\Paragraph{How to choose incremental labeling policies?}
Labeling policies are an important component to our autoregressive framework. 
We demonstrate the level of sensitivity and importance when using a variety of incremental labeling policies. 
Since high value IoU labeling policies only admit \textit{very} well localized boxes as foreground, we refer to the IoU labeling policy of $h\geq0.4$ as lenient, $h\geq0.5$ as moderate, $h\geq0.6$ as strict. 
We train the AR-RPN using labeling techniques of strict-to-lenient, moderate-to-moderate, strict-to-strict, and our primary setting of lenient-to-strict, as shown in Table~\ref{tab:labeling_policy}.
The strict-to-lenient method performs the worse among all settings, degrading by $1.02\%$ MR. 
The moderate-to-moderate performs similarly and degrades by $0.80\%$ MR. 
As shown in Fig.~\ref{phase_visualization}, the primary labeling policy of lenient-to-strict enables the network to start with large clusters of pedestrian box detections and iteratively suppress, resulting in more tight and peaky prediction maps. 
In contrast, strict-to-strict does not \textit{ease} this transition as well resulting in a degradation of $0.42\%$ MR.
We further validate the effect by analyzing the score distributions across \textbf{all} pedestrians in the X/Y directions for the Caltech test dataset, as shown in Fig.~\ref{p_scores}.
We observe a consistent trend in both directions where each successive phase results in a sharper peak with respect to its mean score. 
Each other labeling policy encourages the opposite or encourages the \textit{same} predictions but more accurately. 
On a related point, we furhter examine the disagreements between phases ($\Delta P_{1\to3}$ colored magenta, Fig.~\ref{phase_visualization}) which re-affirms phases logically agree on centroids of pedestrians.
This analysis further shows that most suppression appears to be due to poorly localized boxes primarily in Y-direction (e.g., offset from the legs or head of a pedestrian).

For completeness, we further evaluate the extreme case where there is \textbf{no} incremental supervision or autoregressive flow within the network as included in Table~\ref{tab:labeling_policy}. 
In this case, the core $3$-phase network architecture is kept intact, except the prediction layers and concatenation have been removed from phases $1\to2$ and $2\to3$, therefore there is no incremental labeling policy to be decided. 
In doing so, the detection performance degrades by a considerable $2.14\%$ miss rate, which further suggests that making intermediate predictions with the AR-RPN is a critical component to the classification power of our proposed framework.

%% file: sec_5.tex
\section{Conclusion}

In this work, we present an autoregressive pedestrian detection framework which utilizes a novel stackable de-encoder module with convolutional re-sampling layers.
The proposed AR-Ped framework is able to autoregressively produce and refine both features and classification predictions. 
In consequence, the collective phases approximate an ensemble of increasingly more precise classification decisions and results in an overall improved classifier for pedestrian detection. 
We specifically supervise each phase using increasingly stricter labeling policies such that each phase of the network has similar recall as the last but with tighter and more clusterable prediction maps. 
We provide comprehensive ablation experiments to better understand and support each proposed component of our framework.
We attain new state-of-the-art results on the Caltech dataset throughout many challenging experimental settings and achieve a highly competitive accuracy on the KITTI benchmark.